\documentclass[conference]{IEEEtran}
\IEEEoverridecommandlockouts
\usepackage{cite}
\usepackage{amsmath,amssymb,amsfonts}
\usepackage{algorithmic}
\usepackage{graphicx}
\usepackage{textcomp}
\usepackage{makecell}
\usepackage{xcolor}
\usepackage[utf8]{inputenc}
\usepackage{booktabs}
\usepackage{multirow}
\usepackage{pifont}
\usepackage{url}

\usepackage[table]{xcolor}
\newcommand{\cmark}{\ding{51}} 
\newcommand{\xmark}{\ding{55}} 
\definecolor{highlightgreen}{rgb}{1.0, 1.0, 0.88}
\def\BibTeX{{\rm B\kern-.05em{\sc i\kern-.025em b}\kern-.08em
    T\kern-.1667em\lower.7ex\hbox{E}\kern-.125emX}}
\begin{document}

\title{Fairness-Aware Mixture-of-Experts via Subgroup Reweighting and Gate Regularization\\
}

\author{\IEEEauthorblockN{Sunhee Hwang}
\IEEEauthorblockA{
\textit{Dongyang Mirae University}\\
Seoul, Republic of Korea \\
sunheehwang@dongyang.ac.kr}
}

\maketitle

\begin{abstract}
Deep learning models often produce performance disparities across demographic groups, due to the training data imbalance with respect to sensitive attributes such as gender or age.
To address this problem, existing work has explored fair representation learning, data re-sampling, and adversarial training, which can be broadly categorized into two main approaches.
Single-stage methods typically learn a shared representation for fairness, but often struggle to handle heterogeneous subgroup distributions.
Two-stage methods learn representations separately from the final prediction task, which can lead to misalignment between fairness objectives and downstream predictions.
We identify routing-induced bias, a failure mode in which subgroup imbalance drives the gating network to route subgroups onto a few experts, and propose an end-to-end Mixture-of-Experts (MoE) framework that corrects it.
Specifically, we apply subgroup reweighting to correct data imbalance, and introduce gate entropy regularization to prevent routing from collapsing onto subgroup attributes, keeping expert utilization both balanced and interpretable.
Beyond improving fairness, the routing distribution offers an interpretable view of how subgroups are allocated across experts.
Experimental results demonstrate that the proposed approach improves fairness while maintaining competitive predictive performance.
\end{abstract}

\begin{IEEEkeywords}
fairness, mixture-of-experts, equalized odds, routing-induced bias, bias mitigation
\end{IEEEkeywords}

\section{Introduction}

Deep neural networks have achieved remarkable success across a wide range of visual recognition tasks. In high-stakes applications such as face analysis, however, their predictions are increasingly required to be not only accurate but also fair across demographic groups defined by sensitive attributes such as gender or age~\cite{mehrabi2021survey, wang2023survey,pessach2022review}. In practice, this requirement is often violated: models tend to perform substantially better on some groups than others. A key reason is that when a target label is statistically correlated with a sensitive attribute in the training data, the model can minimize its loss by exploiting this shortcut rather than learning the true decision rule. As a result, the learned representation entangles the target with the sensitive attribute, and the resulting accuracy gap across groups manifests as unfairness at inference time.

A main challenge is that such bias originates not from the model architecture alone, but from the structure of the training data itself. Real-world datasets are rarely balanced: certain combinations of target label and sensitive attribute—for example, a positive label paired with an underrepresented group—appear far less frequently than others. During training, the model is exposed to the majority subgroups far more frequently and optimizes predominantly for them, while underrepresented subgroups receive little gradient signal and are effectively underfit. This imbalance therefore produces disparities not at the level of individual attributes, but across their \textit{intersections} $(y, s)$, where the scarcest subgroups suffer the largest performance drops~\cite{plecko2025fairness,kim2023fair}.

Existing approaches mitigate bias through representation learning, disentanglement, or data-level interventions~\cite{fdvae,fscl,faircl,fairgdms,shasam,hwang2020exploiting}. However, many rely on decoupled or multi-stage pipelines, in which fair representation learning is performed separately from the downstream prediction task. Such formulations do not fully optimize fairness in an end-to-end manner. This can lead to suboptimal alignment between representation learning and the final prediction objective.

Several methods adopt single-stage training that jointly optimizes task performance and fairness~\cite{lnl,mfd,fairvpt}. However, these approaches typically rely on a shared representation space. This makes it difficult to capture heterogeneous subgroup characteristics while maintaining both fairness and accuracy. As a result, a trade-off between bias mitigation and predictive performance is often observed.

These limitations point to a deeper need: a model that can process heterogeneous subgroups through separate computational pathways rather than a single shared representation. Mixture-of-Experts (MoE) provides exactly this capability through its expert routing mechanism. However, we find that MoE introduces a new and previously underexplored failure mode: under subgroup imbalance, the gating network—optimized predominantly on majority subgroups—correlates its routing decisions with sensitive attributes, concentrating each subgroup onto a few experts and leaving others underutilized. We term this \textit{routing-induced bias}. Crucially, this failure is specific to conditional routing and is therefore invisible to existing fairness methods built on a single shared representation. To address it, we propose \textbf{FAMoE}, a fairness-aware MoE framework that diagnoses and corrects routing-induced bias end-to-end. The source code is available at \url{https://github.com/sunhee-hwang/FAMoE}.

\noindent Our main contributions are summarized as follows:
\begin{itemize}
\item We identify \textit{routing-induced bias} as a previously underexplored failure mode in Mixture-of-Experts models, in which subgroup imbalance drives the gating network to correlate routing decisions with sensitive attributes, yielding skewed expert utilization.
\item We propose FAMoE, which corrects routing-induced bias end-to-end by combining subgroup reweighting with a gate entropy regularization that, unlike its conventional use for load balancing, here keeps routing interpretable by preventing experts from being monopolized by specific subgroups.
\item We show that the routing distribution serves as an interpretable diagnostic of subgroup-level behavior, revealing \textit{how} fairness is achieved an inspectability absent in monolithic representation-based methods.
\item We demonstrate consistent improvements in the fairness
-accuracy trade-off across multiple target and sensitive attribute settings on CelebA, alongside more balanced expert utilization.
\end{itemize}

\begin{figure*}[t]
    \centering
    \includegraphics[width=\textwidth]{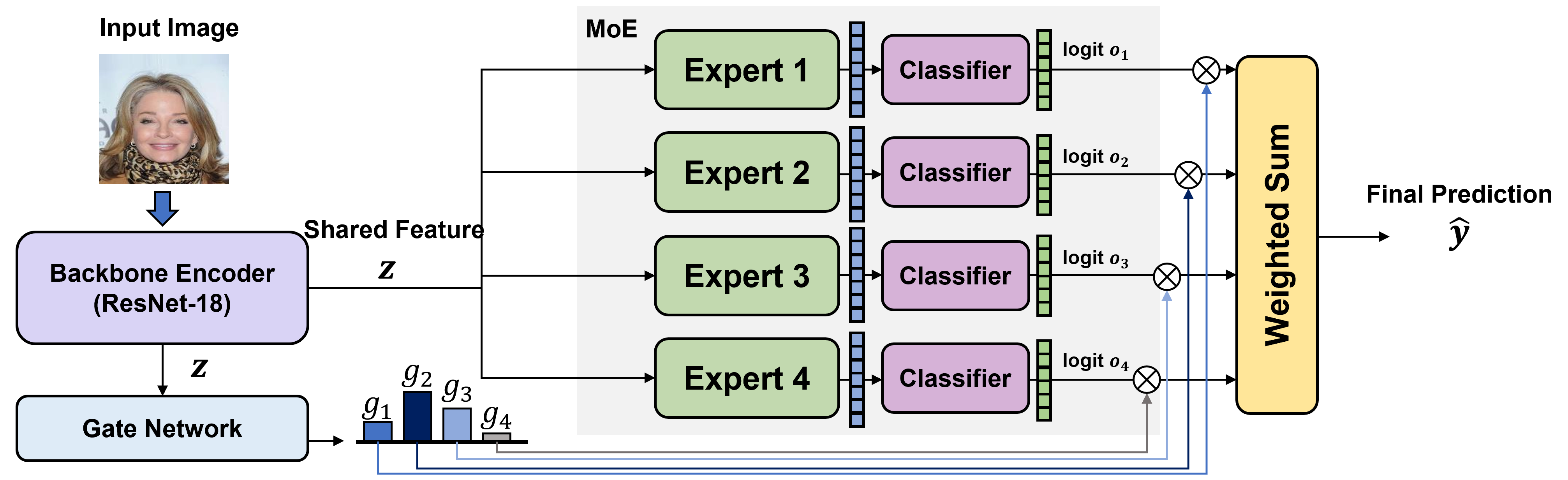}
    \caption{Overview of the proposed FAMoE (Fairness-Aware Mixture-of-Experts) framework.}
    \label{fig:framework}
\end{figure*}

\section{Related Work}

\subsection{MoE and Routing Dynamics}
MoE is a conditional computation framework in which a gating network dynamically assigns inputs to a subset of experts, enabling input-dependent processing and improved model capacity~\cite{moe}. This routing mechanism enables experts to specialize in different regions of the input space. It also facilitates modeling of heterogeneous data. Prior work on MoE has primarily focused on improving efficiency and scalability. Sparse models, exemplified by the Switch Transformer, mitigate computational overhead by selectively activating a subset of experts for each incoming token~\cite{fedus2022switch}. However, imbalanced expert utilization remains a well-known challenge in routing design~\cite{wang2024auxiliary}. Furthermore, empirical evidence consistently demonstrates expert specialization, wherein individual experts encode distinct functional or semantic patterns~\cite{oldfield2024multilinear}.

Despite these advances, existing MoE research has largely emphasized routing as an optimization and efficiency problem. While prior work has considered data heterogeneity and fairness in isolation, how subgroup imbalance shapes routing decisions remains largely unexamined. Recent work has begun to explore fairness in MoE models. For example, FairMOE introduces fairness-aware mechanisms into the expert selection process using counterfactual fairness criteria to promote consistency across protected attributes~\cite{germino2024fairmoe}. However, it does not analyze the interaction between data imbalance and routing dynamics. It also does not control this interaction during training.

\subsection{Learning Fairness Representation}
Bias mitigation in visual classification mainly relies on representation learning. Existing methods adopt either end-to-end joint optimization or multi-stage frameworks to enforce fairness constraints. Under joint optimization settings, adversarial techniques utilize gradient reversal~\cite{grl} to penalize the encoding of sensitive attributes. Mutual information-based objectives~\cite{lnl} operate on a similar principle by minimizing the statistical dependence between learned features and protected variables.

Other strategies address fairness by directly intervening in the feature space. MFD~\cite{mfd} and Fair-VPT~\cite{fairvpt} suppress sensitive cues within a shared representation to prevent the classifier from utilizing demographic information. To formally model the structural interactions between target and sensitive attributes, disentanglement techniques decouple the representations. FD-VAE~\cite{fdvae} partitions the feature space into target-specific, sensitive, and shared components to capture the overlap between features. 

Recent literature extends these representation learning principles to account for inherent data imbalances and spurious correlations. These strategies encompass submodular hard sample mining (SHaSAM~\cite{shasam}), balanced training pair construction via data augmentation (FairCL~\cite{faircl}), and group-aware correlation suppression (Fair-GDMS~\cite{fairgdms}).

Although effective at reducing bias, these methodologies operate exclusively on a globally shared representation space. Processing highly heterogeneous data distributions across varying subgroups within a monolithic feature space restricts the model's representational flexibility. This shared-space bottleneck indicates the necessity of input-dependent computation pathways to accommodate subgroup-specific characteristics. While conditional architectures like Mixture-of-Experts (MoE) provide this structural flexibility, they remain largely underexplored in the context of fairness. Furthermore, as our analysis reveals, standard conditional routing mechanisms are inherently susceptible to subgroup imbalance, necessitating a new framework that jointly addresses data heterogeneity and expert utilization.

\section{Fairness-Aware Mixture-of-Experts Framework}

\subsection{Overall Framework}

We propose a fairness-aware visual classification framework based on a MoE architecture, as illustrated in Fig.~\ref{fig:framework}. Given an input image $x$, a shared backbone encoder first extracts a feature representation $z \in \mathbb{R}^{d}$. This representation is then used as input to both the expert networks and the gating network.

The backbone encoder is shared across all model variants to ensure that differences in performance and fairness can be attributed to the MoE components rather than feature extraction. This design enables a controlled comparison of modeling choices within a unified feature space. The MoE head consists of $N$ experts $\{E_i\}_{i=1}^{N}$ and a gating network $G(\cdot)$. Each expert independently processes the shared feature $z$ and produces an expert-specific output, while the gating network computes a routing weight for each expert based on the same input. The final prediction is obtained by aggregating the expert outputs using the routing weights produced by the gating network. This structure allows the model to adaptively combine multiple expert predictions in a data-dependent manner, enabling flexible modeling of diverse input patterns within a single framework.

\subsection{Expert Prediction and Gated Aggregation}

Given the shared feature representation $z$, each expert $E_i$ produces an intermediate representation, which is passed through a classifier $C_i$ to obtain an expert-specific logit:
\begin{equation}
o_i = C_i(E_i(z)).
\end{equation}

The gating network $G(\cdot)$ computes a routing vector $g = [g_1, \dots, g_N]$, where each element represents the importance of the corresponding expert. The routing weights are normalized using a softmax function:
\begin{equation}
g_i = \frac{\exp(a_i)}{\sum_{j=1}^{N} \exp(a_j)},
\end{equation}
where $a_i$ denotes the routing score for expert $i$.

The final prediction is obtained by aggregating the expert outputs using the routing weights:
\begin{equation}
\hat{o} = \sum_{i=1}^{N} g_i o_i.
\end{equation}
For binary classification, the final prediction probability is computed as $\hat{y} = \sigma(\hat{o})$, where $\sigma(\cdot)$ denotes the sigmoid function.

\subsection{Subgroup Reweighting for Joint Attribute Imbalance}

To address imbalance across target and sensitive attributes, we incorporate subgroup reweighting into the training process. Each sample is associated with a joint subgroup defined by its target label $y$ and sensitive attribute $s$. Let $n_{y,s}$ denote the number of samples in subgroup $(y,s)$. We assign each subgroup $(y,s)$ a weight inversely proportional to its size:
\begin{equation}
w_{y,s} \propto \frac{1}{n_{y,s}}.
\end{equation}
This weighting scheme increases the contribution of underrepresented subgroups during training. We employ a weighted binary cross-entropy loss:
\begin{equation}
\mathcal{L}_{\text{cls}} = \sum_{i=1}^{B} w_{y_i,s_i} \cdot \text{BCE}(\hat{o}_i, y_i),
\end{equation}

where $B$ denotes the batch size.

\begin{table*}[ht]
\centering
\caption{Comparison of fairness–accuracy trade-offs on CelebA across multiple target attributes. We report Equalized Odds (EO), accuracy, and the Fairness–Accuracy Trade-off Score (FATS), under sensitive attributes (Male and Young). A lower FATS indicates a more favorable trade-off. * denotes the method re-implemented by the authors. Male and Young denote the sensitive attribute $s$ used to compute EO; all samples (both $s=0$ and $s=1$) are included in training and evaluation. Bold indicates the best result within each stage category (1-stage and 2-stage), reported separately for each column.}

\label{table:celeba_subset_simplified}
\small
\setlength{\tabcolsep}{3.8pt}
\begin{tabular}{l c c ccc ccc c ccc ccc}
\toprule
\multirow{3}{*}{Method} & \multirow{3}{*}{Year} & \multirow{3}{*}{Stage}
& \multicolumn{6}{c}{Target: Attractiveness} 
& \multicolumn{6}{c}{Target: Big Nose} \\
\cmidrule(lr){4-9} \cmidrule(lr){10-15}
& & 
& \multicolumn{3}{c}{Male} 
& \multicolumn{3}{c}{Young}
& \multicolumn{3}{c}{Male} 
& \multicolumn{3}{c}{Young} \\
& & 
& EO & ACC & FATS
& EO & ACC & FATS
& EO & ACC & FATS
& EO & ACC & FATS \\
\midrule 

ResNet-18~\cite{resnet18} 
& 2016 & \multirow{2}{*}{Baseline}
& 27.8 & 79.6 & 27.9
& 16.8 & 79.8 & 16.9
& 17.6 & 84.0 & 17.7
& 14.7 & 84.5 & 14.8 \\

*ResNet-18~\cite{resnet18} + MoE~\cite{moe} 
& 2016 / 1991 & 
& 13.3 & 81.0 & 13.4
& 22.4 & 81.1 & 22.5
& 16.0 & 84.1 & 16.1
& 14.1 & 84.3 & 14.2 \\

\midrule

FD-VAE~\cite{fdvae}
& 2021 & \multirow{5}{*}{2-stage}
& 15.1 & 76.9 & 15.2
& 14.8 & 77.5 & 14.9
& 11.2 & 81.6 & 11.3
& 6.7 & 81.7 & 6.8 \\

FSCL~\cite{fscl}
& 2022 &
& 6.5 & 79.1 & 6.6
& 12.4 & 79.1 & 12.5
& 4.7 & 82.9 & 4.8
& 4.8 & 84.1 & 4.9 \\

FairCL~\cite{faircl}
& 2023 &
& 16.8 & 75.3 & 16.9
& 13.1 & 76.9 & 13.2
& 8.4 & 80.1 & 8.5
& 9.2 & 80.3 & 9.3 \\

Fair-GDMS~\cite{fairgdms}
& 2026 &
& 7.0 & \textbf{83.2} & 7.1
& - & - & -
& 13.2 & \textbf{84.8} & 13.3
& - & - & - \\

SHaSAM~\cite{shasam}
& 2026 &
& \textbf{5.5} & 81.3 & \textbf{5.6}
& \textbf{9.9} & \textbf{79.6} & \textbf{10.0}
& \textbf{3.3} & 84.7 & \textbf{3.4}
& \textbf{3.9} & \textbf{87.0} & \textbf{4.0} \\

\midrule

GRL~\cite{grl}
& 2018 & \multirow{5}{*}{1-stage}
& 24.9 & 77.2 & 25.0
& 14.7 & 74.6 & 14.8
& 14.0 & \textbf{82.5} & 14.1
& 10.0 & \textbf{83.3} & 10.1 \\

LNL~\cite{lnl}
& 2019 &
& 21.8 & 79.9 & 21.9
& 13.7 & 74.3 & 13.8
& 10.7 & 82.3 & 10.8
& 6.8 & 82.3 & 6.9 \\

MFD~\cite{mfd}
& 2021 &
& 7.4 & 78.0 & 7.5
& 14.9 & \textbf{80.0} & 15.0
& 7.3 & 78.0 & 7.4
& 5.4 & 78.0 & 5.5 \\

Fair-VPT~\cite{fairvpt}
& 2024 &
& 12.0 & 78.6 & 12.1
& - & - & -
& 15.9 & 79.9 & 16.0
& - & - & - \\

\rowcolor{highlightgreen}
\textbf{FAMoE (Ours)} 
& 2026 &
& \textbf{4.1} & \textbf{80.0} & \textbf{4.2}
& \textbf{8.3} & 79.8 & \textbf{8.4}
& \textbf{4.5} & 80.1 & \textbf{4.6}
& \textbf{3.3} & 80.8 & \textbf{3.4} \\

\bottomrule
\end{tabular}
\end{table*}

\subsection{Gate Entropy Regularization}

To promote balanced utilization of experts, we introduce a gate entropy regularization term. Given the routing distribution $g$ for each input, the entropy is defined as:
\begin{equation}
H(g) = -\sum_{i=1}^{N} g_i \log g_i.
\end{equation}

We add this term to encourage higher entropy in the routing distribution. The overall loss function is defined as:
\begin{equation}
\mathcal{L} = \mathcal{L}_{\text{cls}} - \lambda_{\text{ent}} H(g),
\end{equation}
where $\lambda_{\text{ent}}$ controls the strength of the regularization. This regularization reduces overly concentrated routing decisions. It also encourages the utilization of multiple experts during training. As a result, the model maintains diverse expert contributions while performing data-dependent aggregation. Importantly, this regularization is not merely a fairness penalty but a mechanism that keeps the routing distribution interpretable: by preventing an expert from being monopolized by a specific subgroup, the resulting gate weights remain a faithful diagnostic of how the model distributes subgroups across experts.

\subsection{Training Objective}

The overall training objective combines subgroup reweighting and gate entropy regularization. Different model variants are instantiated by selectively enabling each component. Specifically, the standard MoE model is obtained by setting $w_{y,s} = 1$ and $\lambda_{\text{ent}} = 0$, while the reweighted model incorporates subgroup weights with $\lambda_{\text{ent}} = 0$. The full model includes both subgroup reweighting and entropy regularization. All models are trained end-to-end using stochastic gradient descent, where the encoder, experts, and gating network are jointly optimized.

\section{Experiments}

\subsection{Experimental Setup}

We evaluate the proposed method on the CelebA dataset, which contains facial images annotated with multiple binary attributes. Following standard practice, we consider binary classification tasks with a designated target attribute and a sensitive attribute. Each sample is associated with a pair $(y, s)$, where $y$ denotes the target label and $s$ denotes the sensitive attribute. The dataset is split into training, validation, and test sets according to the official partition. Images are resized to $128 \times 128$ and normalized. We use standard data augmentation, including random resized cropping and horizontal flipping during training. All models share the same backbone encoder based on ResNet-18. The MoE head consists of four experts and a gating network that produces a routing distribution over experts. 
Each expert is a two-layer MLP ($512 \rightarrow 128 \rightarrow 64$) with ReLU activation, followed by an expert-specific linear classifier ($64 \rightarrow 1$). 
All experts share the same architecture but maintain independent parameters. The gating network is a single linear layer ($512 \rightarrow 4$) followed by a softmax.
Models are trained using the Adam optimizer with a learning rate of $10^{-4}$ and weight decay of $10^{-4}$ for 30 epochs. 

We evaluate performance using classification accuracy, Equalized Odds (EO)~\cite{hardt2016equality}, and the Fairness–Accuracy Trade-off Score (FATS). EO measures the difference in prediction performance across subgroups defined by $(y, s)$. Following~\cite{fscl}, we compute EO as the average accuracy gap across sensitive groups within each target class. FATS is defined as:
\begin{equation}
\text{FATS} = \text{EO} + \alpha (1 - \text{Acc}),
\end{equation}
where $\alpha$ controls the trade-off between fairness and accuracy. In our experiments, we set $\alpha = 0.5$ to weight fairness and accuracy degradation equally, and a lower FATS indicates a more favorable balance between fairness and predictive performance. The best model is selected based on the validation FATS.

\subsection{Main Results}
Table~\ref{table:celeba_subset_simplified} presents the comparison with existing fairness-aware methods across different target and sensitive attribute settings. Compared to one-stage methods such as GRL, LNL, and MFD, the proposed method consistently improves fairness while avoiding large drops in accuracy. Compared to two-stage approaches, including FD-VAE, FairCL, and Fair-GDMS, the proposed method achieves a more favorable balance between fairness and accuracy, as reflected in consistently lower FATS values. These results indicate that jointly addressing subgroup imbalance and routing decisions leads to improved fairness compared to existing approaches.

\begin{table}[t]
\centering
\caption{Ablation study of proposed components on CelebA (Target: Attractiveness).}
\label{table:ablation_attractive}
\small
\setlength{\tabcolsep}{4pt}
\begin{tabular}{cccccc}
\toprule
& \multicolumn{2}{c}{Components}
& \multicolumn{3}{c}{Sensitive: Male} \\
\cmidrule(lr){2-3} \cmidrule(lr){4-6}
& \makecell{Subgroup \\ Reweighting}
& \makecell{Gate Entropy \\ Regularization}
& EO & ACC & FATS \\
\midrule

& \xmark & \xmark 
& 13.3 & 81.0 & 13.4 \\

& \cmark & \xmark 
& 4.2 & 79.6 & 4.3 \\

\rowcolor{highlightgreen}
& \cmark & \cmark 
& \textbf{4.1} & 80.0 & \textbf{4.2} \\

\bottomrule
\end{tabular}
\end{table}

\subsection{Ablation Study}

\begin{figure}[t]
    \centering
    \includegraphics[width=\linewidth]{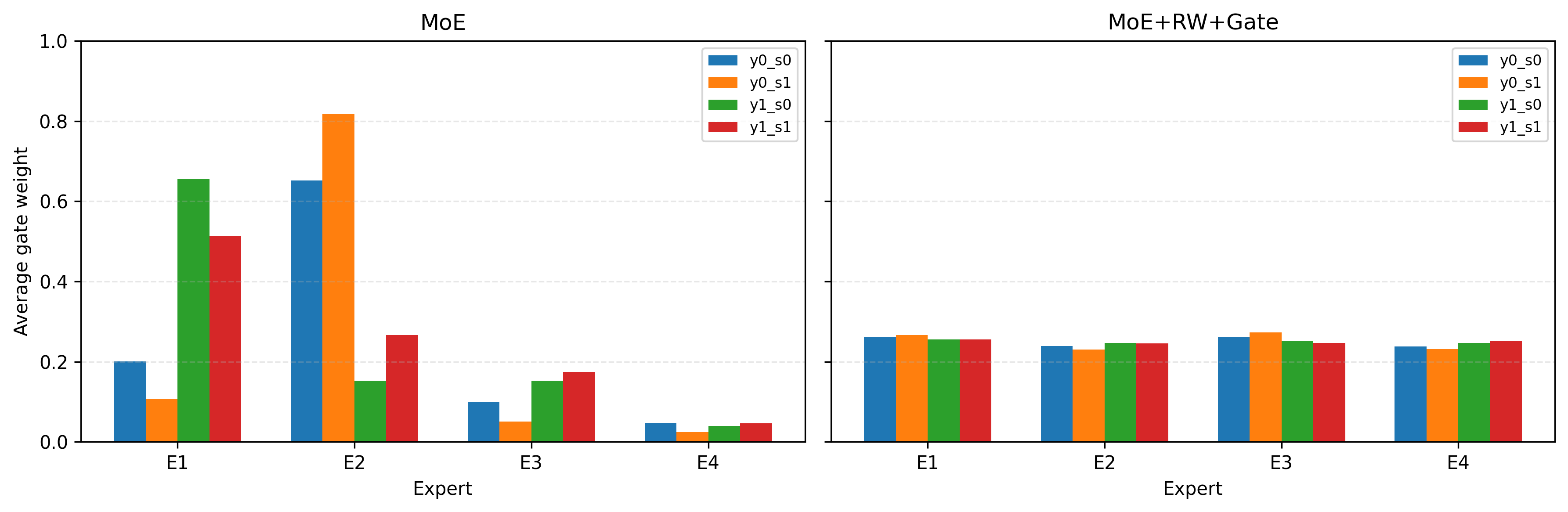}
    \caption{Average gate weights across subgroups for the standard MoE and the proposed method.}
    \label{fig:gate_analysis}
\end{figure}

Table~\ref{table:ablation_attractive} shows the effect of each component in the proposed framework. The standard MoE model achieves an EO of 13.3, indicating that MoE alone does not improve fairness. Adding subgroup reweighting reduces EO from 13.3 to 4.2, confirming that data imbalance is a primary driver of routing bias. Reweighting alone, however, leaves routing concentration uncontrolled; adding gate entropy regularization further improves EO to 4.1, recovers accuracy, and yields the balanced, interpretable routing analyzed in Fig.~\ref{fig:gate_analysis}. Entropy regularization thus complements reweighting by promoting balanced expert utilization.

\subsection{Analysis of Expert Utilization}

To investigate expert utilization patterns, we analyze the average gate weights across demographic subgroups (Fig.~\ref{fig:gate_analysis}). In the standard MoE baseline, routing decisions exhibit pronounced skewness. Specific experts are disproportionately activated by certain subgroups, while others remain underutilized. This observation indicates that the baseline routing mechanism inadvertently captures spurious correlations with sensitive attributes. In contrast, our method effectively mitigates this dependency, yielding a more balanced distribution of routing weights. These empirical findings demonstrate that our approach prevents a single subgroup from dominating expert capacity, thereby promoting more equitable expert utilization across subgroups. This analysis is only possible because routing exposes expert allocation explicitly. Monolithic models that share a single representation space may reach comparable fairness scores, yet they offer no comparable window into how fairness is achieved. The gate-weight distribution in Fig.~\ref{fig:gate_analysis} thus serves as a built-in diagnostic of subgroup-level model behavior, a form of routing-level interpretability not available in monolithic representation-based methods.

\section{Conclusion}
This study addresses the degradation of fairness in MoE models caused by imbalanced data distributions. We show that, under subgroup imbalance, conditional routing tends to induce systematic bias, skewing expert utilization based on subgroup attributes. Recognizing that existing representation-focused methods fail to capture this routing-induced bias, we propose a framework combining subgroup reweighting with gate entropy regularization. Our empirical results demonstrate that regularizing routing behaviors while correcting data imbalance promotes more balanced expert assignment and substantially improves the fairness–accuracy trade-off. Overall, this work highlights the critical necessity of explicitly accounting for expert utilization, which not only improves fairness but also renders the model's subgroup-level behavior interpretable through its routing distribution—a form of transparency not offered by monolithic fairness methods.

\section{Acknowledgment}
This work was supported by the Korea Foundation for Women in Science, Engineering and Technology (WISET) with funding from the Science and Technology Promotion Fund and Lottery Fund (WISET 2026-236).

\bibliographystyle{IEEEtran} 
\bibliography{references}

\end{document}